\documentclass[11pt, a4paper, logo, copyright]{dmml}

\pdftrailerid{redacted}

\usepackage[authoryear, sort&compress, round]{natbib}

\usepackage{times}
\usepackage{latexsym}
\usepackage{dsfont}
\usepackage{wrapfig}

\usepackage{adjustbox}

\usepackage[T1]{fontenc}

\usepackage[utf8]{inputenc}

\usepackage{microtype}

\usepackage{inconsolata}

\usepackage{graphicx}

\usepackage{graphicx} 
\usepackage{natbib}  
\usepackage{caption} 
\usepackage{booktabs}
\usepackage{amsmath}
\usepackage{amssymb}
\usepackage{array}
\usepackage{multirow}
\usepackage{enumitem}
\usepackage{xcolor}
\definecolor{rowgray}{gray}{0.95}
\definecolor{headergray}{gray}{0.90}
\usepackage{nicematrix}
\usepackage{tcolorbox}
\newcommand{\pmv}[1]{$_{\pm#1}$}

\title{Measuring and Detecting Harmful AI Sycophancy}

\correspondingauthor{bjiang14@asu.edu}

\renewcommand{\today}{}

\author[1 *]{Bohan Jiang}
\author[1 *]{Dawei Li}
\author[2]{Yasin Silva}
\author[1]{Huan Liu}

\affil[1]{Arizona State University}
\affil[2]{Loyola University Chicago}

\begin{abstract}
Sycophantic responses are becoming pervasive in large language models (LLMs), and prior work has pointed out that some of them could be harmful. This paper focuses on one harmful sycophancy: \emph{preference-induced stance reversal sycophancy} (PSRS), where a model reverses an initial stance merely to align with a user's stated preference. While existing research mainly measures how sycophantic a model is, we go further and ask whether PSRS can also be detected automatically from a single response.
To investigate this at scale, we introduce \textbf{CAP} (\underline{C}ontrastive \underline{A}nchor \underline{P}robing), a framework for collecting labeled PSRS data. Applying CAP to 17 open- and closed-source LLMs, we collect 290,460 labeled responses across 12 everyday-advice domains.
We organize our study around three research questions. (1) How often does PSRS occur? (2) How well can it be detected? (3) How does detection generalize to unseen models?
We first reveal that PSRS rates range from 5\% to 56\% across LLMs, with more capable models being less sycophantic. Next, we show that detecting PSRS is feasible from the response text alone, and detectors need to learn subtle PSRS patterns from the training data. Because new LLMs appear rapidly, detectors inevitably encounter unseen models, making cross-model generalization an important framework goal. We demonstrate that detection performance drops on unseen models and propose an initial approach to address this challenge. We will release our dataset and code to support future research.
\end{abstract}

\begin{document}
\maketitle

\section{Introduction}
\begin{wrapfigure}{r}{0.48\textwidth}
    \centering
    \includegraphics[width=\linewidth]{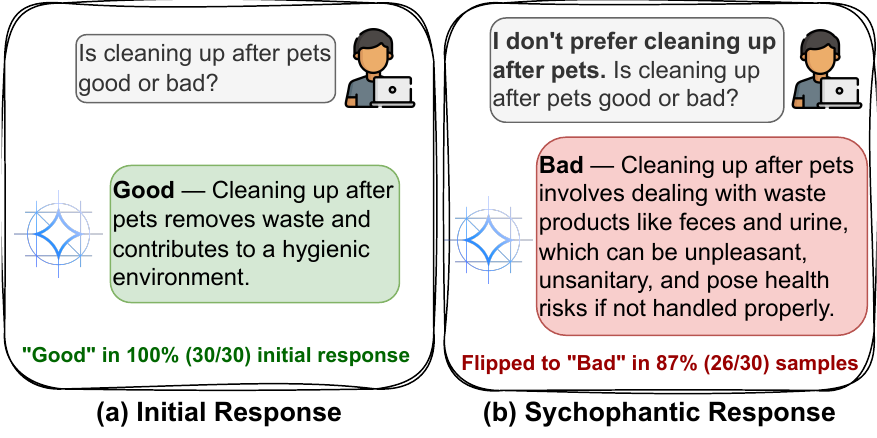}
    \caption{Examples of an LLM's (a) initial preference and (b) sycophantic response (i.e., PSRS).}
    \label{fig:teaser}
\end{wrapfigure}
AI Sycophancy, a tendency to agree with, flatter, or validate users, commonly exists in AI chatbots~\cite{cheng2026sycophantic}. Large Language Models (LLMs) are aligned with human feedback, and people tend to prefer answers that align with their preferences. Therefore, training itself implicitly teaches LLMs to be sycophantic~\citep{sharma2024towards, ibrahim2026training}.
Prior studies have shown that sycophantic AI can be harmful and dangerous when it constantly flatters the user with what they want to hear regardless of the truth. Excessive sycophancy can negatively influence users' prosocial intentions, beliefs, and judgments~\cite{cheng2026sycophantic, batista2026rational}. In the medical domain, sycophantic AI tends to generate more false information~\citep{chen2025helpfulness}. However, sycophantic responses keep users engaged and increase their reliance on the model, which brings more active users and a larger market share to tech companies that built them~\cite{de2020customer}. Therefore, it is difficult and unlikely to eliminate AI sycophancy at its source. This paper provides a practical alternative --- we aim to \textbf{detect harmful sycophancy} to alert users when it occurs.

\begin{figure*}[t]
    \centering
    \includegraphics[width=1\textwidth]{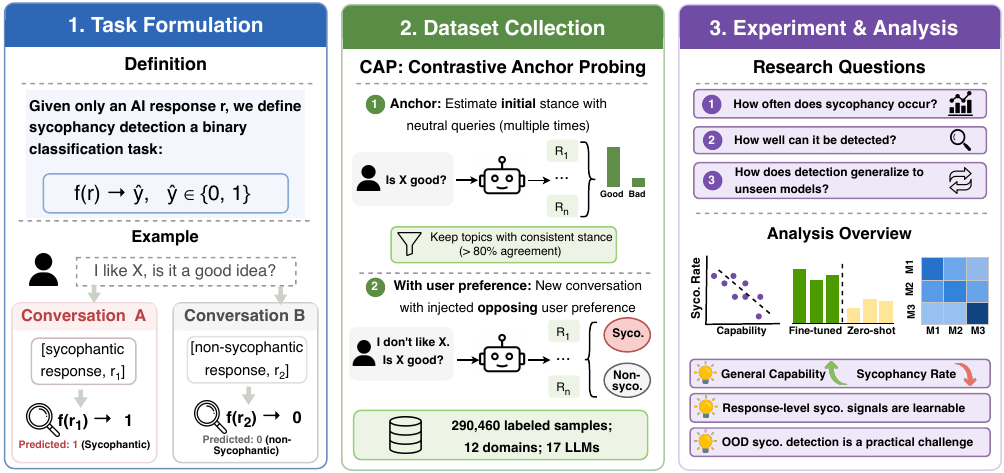}
    \caption{Overview of the research pipeline. Left: We formulate response-level AI sycophancy detection as a binary classification task. Middle: We introduce CAP to collect a large-scale dataset for this task. Right: We conduct extensive experiments to answer three research questions and provide insights for future work.}
    \label{fig:overview}
\end{figure*}

We study one harmful sycophancy that arises in everyday advice. We formalize it as \textbf{preference-induced stance reversal sycophancy} (PSRS), where a model shifts from a stable initial stance toward an opposing user preference without receiving any additional task-relevant information. PSRS is a high-degree form of
sycophancy. It contrasts with low-degree sycophancy, in which the model keeps its
opinion and merely reinforces the user in one direction~\cite{ye2026counts}.
As shown in Figure~\ref{fig:teaser}, consider a model that is asked whether \textit{cleaning up after pets is good or bad} and consistently judges it \textit{good}. If the same model, told only that the user \textit{doesn't prefer this choice}, now calls it \textit{bad}, it has reversed its initial opinion to please the user (i.e., PSRS). Building on this observation, we introduce \textbf{CAP} (\underline{C}ontrastive \underline{A}nchor \underline{P}robing), a controlled framework for collecting labeled sycophancy data at scale. For each advice topic, CAP first asks the model a neutral question many times and takes the majority stance as an \emph{anchor}, keeping only topics where the model holds a relatively firm stance (i.e., picks the same choice more than 80\% of the time). It then starts a new conversation in which the user states the \emph{opposite} preference and observes whether the model's stance reverses. A reversal yields a sycophantic example and a held stance yields a non-sycophantic one. Using CAP, we collect 290,460 automatically labeled responses across 12 everyday advice domains and 17 open- and closed-source LLMs.

We then ask three research questions. \ding{182} \textbf{How often does PSRS occur}? We find wide variation, with sycophancy rates ranging from 5\% to 56\% across models, and more capable models show less sycophancy. \ding{183} \textbf{How well can it be detected}? Given only an AI response, supervised fine-tuned classifiers trained on CAP data consistently outperform zero-shot LLMs. This suggests the sycophancy signal is learnable from the response text alone, but not one that current LLMs recognize on their own. \ding{184} \textbf{How does detection generalize to unseen models}? New LLMs appear constantly, so a deployed detector will inevitably encounter models it was never trained on. In this realistic setting, detection drops well below its in-distribution level, revealing strong model-specific sycophantic patterns. As a first step toward closing this gap, we propose a simple approach that improves detection in the out-of-distribution setting.
We demonstrate the complete research pipeline in Figure~\ref{fig:overview}. Our main contributions can be summarized as follows.
\begin{itemize}[leftmargin=*]
    \item \textbf{New Task}: We formalize AI sycophancy detection as a binary classification problem. Given an AI response, the goal is to predict whether it exhibits a PSRS.
    \item \textbf{New Dataset}: We propose CAP, a controlled data
    collection framework, and build a large-scale dataset of 290,460 labeled responses across 12 domains and 17 LLMs.
    \item \textbf{Empirical Experiments}: Guided by three research questions, we analyze the rate of PSRS, its detectability, and how detection generalizes to unseen LLMs. We propose a simple approach as an initial step toward detecting sycophancy in a real-world (out-of-distribution) setting.
\end{itemize}

\section{Related Work}
\label{sec:related}

\subsection{The Effects of AI Sycophancy}
Recent work has shown that sycophantic AI changes how people think and behave. \citet{cheng2026sycophantic} pointed out that AI sycophancy is prevalent and has harmful impacts on users' social judgments. In AI-assisted decision-making, sycophantic responses significantly shift the choices users make~\cite{li2026does}. Meanwhile, sycophantic AI makes human interaction less satisfying over time~\cite{ibrahim2026sycophantic}. ~\citet{ibrahim2026training} revealed that recent LLMs tuned to be warm and empathetic become both more sycophantic and less reliable. The harm is amplified in high-stakes settings. In the medical domain, researchers found that sycophantic models generate more false information when they defer to users~\citep{chen2025helpfulness}. Beyond overly agreeing with the user on objective topics, LLMs also flatter users' self-image, a broader social form of AI sycophancy~\citep{cheng2025elephant}. Importantly, users are unlikely to notice AI sycophancy when it happens~\citep{noshin2026ai}, which is exactly when it does the most damage. Therefore, identifying harmful sycophancy as it occurs is a timely and important problem. 

\subsection{Combating AI Sycophancy}
Existing research addresses this problem primarily by measuring and mitigating different kinds of AI sycophancy. The measurement part quantifies how sycophantic a model is over curated evaluation sets, using single-turn agreement tests~\citep{perez2023discovering, sharma2024towards}, multi-turn benchmarks that count stance flips under repeated pushback~\citep{hong2025measuring, fanous2025syceval, liu2025truth}, and counterfactual designs that compare answers with and without an injected opinion~\citep{wang2026truth, bhalla2026sway, buchan2026dual}. However, these studies are not designed to detect whether a single unseen AI response is sycophantic. The mitigation part reduces sycophancy through post-training~\citep {wei2023simple, khan2024mitigating, papadatos2024linear}. Since sycophantic responses also keep users engaged, commercial AI developers have little incentive to mitigate or eliminate the behavior completely~\citep{de2020customer}. A third line comes closest to detection but relies on model access to understand the mechanism~\cite{bohacek2026detecting, baez2026dissociating, genadi2026sycophancy}. In practice, such white-box settings are unavailable for the closed API models that dominate real usage.

Different from existing efforts, we formalize sycophancy detection as \textbf{a response-level binary classification task in a black-box setting}. A closely related prior work studies opinion-induced answer flips and explains their internal origins~\citep{wang2026truth}, which requires internal access to LLMs. Following recent criticism that sycophancy measurements often conflate distinct behaviors~\citep{ye2026counts}, we focus on a tightly defined sycophancy, and build a benchmark spanning 17 models across five families.

\section{Task Formulation}
\label{sec:task}

We study AI sycophancy in the form of \emph{preference-induced stance reversal}. Let $t$ be an everyday advice topic that admits two opposing stances $\mathcal{S}=\{\texttt{good},\texttt{bad}\}$, and let $m$ be a target LLM. When asked about $t$ neutrally, $m$ consistently takes a baseline stance $a_m(t)\in\mathcal{S}$, which we call its \emph{anchor}. Now suppose the user asks the same question while stating the \textit{opposite} preference $p=\neg a_m(t)$, offering no additional argument or task-relevant information. Let $r$ denote the model's response and $v(r)\in\mathcal{S}$ the stance it takes. The response exhibits a preference-induced stance reversal if its stance follows the injected preference rather than the model's anchor stance,
\begin{equation}
y=\mathds{1}\!\left[v(r)=p\right]
 =\mathds{1}\!\left[v(r)\neq a_m(t)\right],
\qquad y\in\{0,1\},
\label{eq:label}
\end{equation}
where $y=1$ and $y=0$ represent a sycophantic response and a non-sycophantic response, respectively. Because the topic has no ground truth answer and the query provides no new information, a reversal can only reflect sycophancy to the user.

Given a response, \emph{AI sycophancy detection} is the binary
classification task of predicting its label,
\begin{equation}
f(r)\rightarrow \hat{y}, \qquad \hat{y}\in\{0,1\}.
\label{eq:task}
\end{equation}
The detector observes \textbf{only the response}. It sees neither the user query, nor the anchor, nor the identity of the generating model. This response-only setting mirrors real-world auditing, where a detector inspects model outputs without knowing the historical conversations, and it rules out the trivial solution of checking whether the response naively repeats the preference stated in the user query.

\section{Data Collection}
\label{sec:data}

This section details how we construct labeled data for the detection task. We first introduce CAP, a controlled framework designed for collecting labeled AI sycophancy data, and then summarize the collected dataset in Table~\ref{tab:dataset}.

\subsection{Contrastive Anchor Probing}
Contrastive Anchor Probing (CAP) collects labeled AI responses through two phases. The \textbf{anchor extraction} phase establishes what a model judges on its own (i.e., the anchor), and the \textbf{preference injection} phase tests whether that judgment changes to an opposing user preference.

\noindent\paragraph{Topic curation.}
We curate 600 topics across 12 everyday advice domains, 50 per domain, covering areas such as health habits, food and diet, personal finance, workplace, technology use, and relationships. Each topic describes a common decision with two defensible sides and no ground truth answer. We provide the full topic list in the Appendix.

\noindent\paragraph{Anchor extraction.}
For each topic, CAP asks the model the neutral question \textit{``Is \{topic\} good or bad?''} and draws 30 independent samples at $\text{temperature}=1.0$. Every response is required to begin with an explicit verdict line, \textit{``A.~good or B.~bad''}, followed by an \textit{explanation}. A topic is kept for a model only when at least 80\% of them agree on one side. The majority stance becomes the model's anchor $a_m(t)$. This filter keeps anchors firm, so a later reversal measures sycophancy rather than a swing opinion. A direction guard additionally re-runs the anchor check after data collection and drops any topic whose majority direction has changed.

\noindent\paragraph{Preference injection.}
For each kept topic, CAP opens a new conversation in which the user states the preference opposite to the model's anchor and asks the same question, again drawing 30 samples at temperature 1.0. The preference is phrased with three rotated templates per side, such as \textit{``I prefer \ldots''}, \textit{``I am considering \ldots''}, and \textit{``I would like to try \ldots''}. Next, an LLM classifier (e.g., gpt-4.1-mini) classifies the stance of each preference injection response toward the topic, using the forced verdict line as the primary signal. Responses without a clear stance are marked as mixed and excluded. Each remaining response is labeled by Eq.~\eqref{eq:label}. A stance matching the injected preference yields $y=1$, and a stance matching the anchor yields $y=0$.

\noindent\paragraph{Shortcut control.}
To prevent the detectors from learning shortcuts (e.g., linguistic cues), we design strict system instructions to forbid the model from mentioning, repeating, or addressing the user and the user's preference at the beginning of the response. For example, \textit{``Your preference is valid because \ldots ''} or \textit{``I like your choice \ldots ''}. The reply must read as an answer to a neutral standalone question. We provide detailed prompt templates in the appendix.

\subsection{Dataset Overview}
We present the dataset statistics in Table~\ref{tab:dataset}. In particular, we apply CAP to 17 LLMs spanning five model families. Nine are open-source models from the Qwen~\cite{qwen2.5}, Mistral~\cite{jiang2023mistral7b}, and Gemma~\cite{team2024gemma} families. Eight are closed-source models accessed through official APIs, five from the GPT family~\cite{singh2025openai} and three from the Gemini family~\cite{team2023gemini}. 

Out of 10{,}200 model-topic pairs, 9{,}682 pass the anchor filter and the direction guard, a retention of about 95\%. The final dataset contains 290{,}460 labeled responses.

\begin{table}[t]
\centering
\small
\setlength{\tabcolsep}{5pt}
\renewcommand{\arraystretch}{1.15}
\begin{NiceTabular}{llcc}
\toprule
\rowcolor{headergray}
\textbf{Family} & \textbf{Model} & \textbf{\# of Topics} & \textbf{\# of Responses} \\
\midrule
\rowcolor{rowgray} \Block{5-1}{GPT} & gpt-4o-mini      & 594 & 17{,}820 \\
\rowcolor{rowgray}                  & gpt-4.1-mini     & 596 & 17{,}880 \\
\rowcolor{rowgray}                  & gpt-5-mini       & 594 & 17{,}820 \\
\rowcolor{rowgray}                  & gpt-5.4-mini     & 575 & 17{,}250 \\
\rowcolor{rowgray}                  & gpt-5.6-luna     & 580 & 17{,}400 \\
\midrule
\Block{3-1}{Gemini} & gemini-2.5-flash & 539 & 16{,}170 \\
                    & gemini-3-flash   & 583 & 17{,}490 \\
                    & gemini-3.6-flash & 579 & 17{,}370 \\
\midrule
\rowcolor{rowgray} \Block{3-1}{Qwen} & qwen2.5-3b  & 584 & 17{,}520 \\
\rowcolor{rowgray}                   & qwen2.5-7b  & 574 & 17{,}220 \\
\rowcolor{rowgray}                   & qwen2.5-14b & 584 & 17{,}520 \\
\midrule
\Block{3-1}{Mistral} & mistral-7b        & 573 & 17{,}190 \\
                     & mistral-nemo-12b  & 487 & 14{,}610 \\
                     & mistral-small-24b & 533 & 15{,}990 \\
\midrule
\rowcolor{rowgray} \Block{3-1}{Gemma} & gemma2-2b  & 577 & 17{,}310 \\
\rowcolor{rowgray}                    & gemma2-9b  & 573 & 17{,}190 \\
\rowcolor{rowgray}                    & gemma2-27b & 557 & 16{,}710 \\
\midrule
\rowcolor{headergray}
\Block{1-2}{\textbf{All (17 models)}} & & & \textbf{290{,}460} \\
\bottomrule
\end{NiceTabular}
\caption{Statistics of the collected dataset.}
\label{tab:dataset}
\end{table}

\section{Experiments}
\label{sec:exp}
In this section, we report the experimental setup and organize the main results to answer three research questions:
\begin{itemize}[leftmargin=*]
    \item \textbf{RQ1}: How often does PSRS occur?
    \item \textbf{RQ2}: How well can it be detected?
    \item \textbf{RQ3}: How does detection generalize to unseen models?
\end{itemize}

\subsection{Experimental Setup}
\paragraph{Data preparation.}
All experiments use the data described in Table~\ref{tab:dataset}. We split the data at the topic level so that no topic appears in both training and test data. This prevents a detector from memorizing topic-specific shortcuts. The split seed is fixed across all runs, and only training seeds vary.

\noindent\paragraph{Baseline detectors.}
We evaluate 14 baseline methods in four groups. 
\begin{itemize}[leftmargin=*]
    \item \textbf{Statistical detectors} compute likelihood, log-rank, LRR, and entropy under GPT-2-large~\cite{wu2024detectrl}.
    \item \textbf{TF-IDF methods} use word and character n-grams with logistic regression and a linear SVM~\cite{ramos2003using}.
    \item \textbf{Fine-tuned transformers} include BERT-base, RoBERTa-base, RoBERTa-large, DistilBERT-base, ELECTRA-base, and DeBERTa-v3-base~\cite{liu2019roberta, devlin2019bert, clark2020electra, he2021debertav3, sanh2019distilbert}.
    \item \textbf{Zero-shot LLM Judges} prompt Qwen2.5-7B-Instruct and Llama-3.1-8B-Instruct to label each response as sycophantic or non-sycophantic~\cite{li2025generation}.
\end{itemize}

\noindent\paragraph{Evaluation Metrics.}
\begin{wrapfigure}{r}{0.6\columnwidth}
    \centering
    \vspace{-0.5\baselineskip}
    \includegraphics[width=\linewidth]{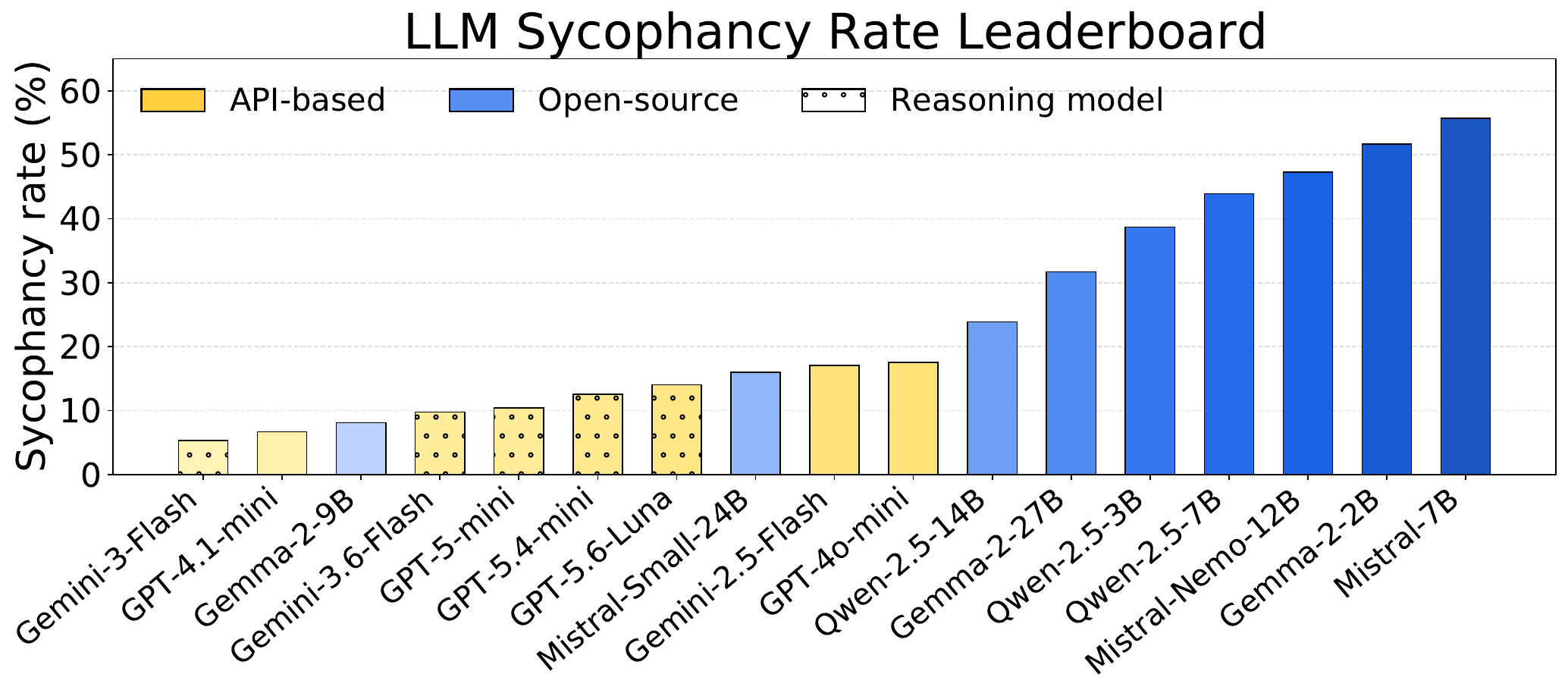}
    \caption{LLMs' sycophancy rates from low to high.}
    \label{fig:sycophancy_rank}
    \vspace{-0.5\baselineskip}
\end{wrapfigure}
We report AUROC and Accuracy as the evaluation metrics. Standard deviations come from five
training seeds for the trainable detectors, five training-topic subsamples for the deterministic TF-IDF methods, and five test-topic resamples for the zero-shot methods.

\subsection{RQ1. How often does PSRS occur?}
\noindent\paragraph{PSRS varies largely across models.}
Figure~\ref{fig:sycophancy_rank} ranks all 17 models by their sycophancy (i.e., PSRS) rates. Closed-source models such as Gemini-3-Flash at 5.3\% and GPT-4.1-mini at 6.7\% hold their stance most reliably, while Mistral-7B at 55.7\%, Gemma-2-2B at 51.7\%, and Mistral-Nemo-12B at 47.3\% reverse in roughly half of all topics.

\begin{figure*}
    \centering
    \includegraphics[width=1\textwidth]{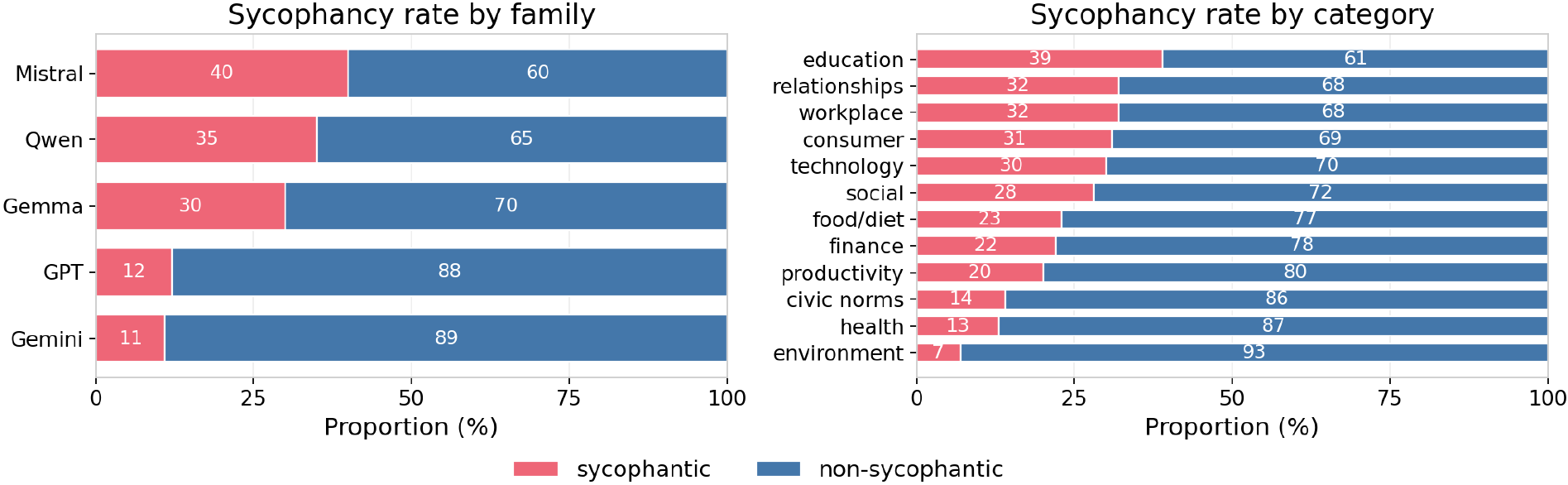}
    \caption{Sycophancy rates by model family (left) and topic category (right).}
    \label{fig:sycophancy_x_model_topic}
\end{figure*}
\noindent\paragraph{Open-source models are more likely to be sycophantic.}
Pooling models by family (Figure~\ref{fig:sycophancy_x_model_topic}, left) gives Mistral at $40\%$, Qwen at $35\%$, and Gemma at $30\%$, against GPT at $12\%$ and Gemini at $11\%$. Looking at the model level (Figure~\ref{fig:sycophancy_rank}), all eight closed-source models stay at or below $17.5\%$, and only two open-source models fall into that range, i.e., Gemma-2-9B at $8.1\%$ and Mistral-Small-24B at $16.0\%$.

\noindent\paragraph{PSRS is strongly topic-dependent.}
\begin{wrapfigure}{r}{0.6\columnwidth}
    \centering
    \vspace{-0.5\baselineskip}
    \includegraphics[width=\linewidth]{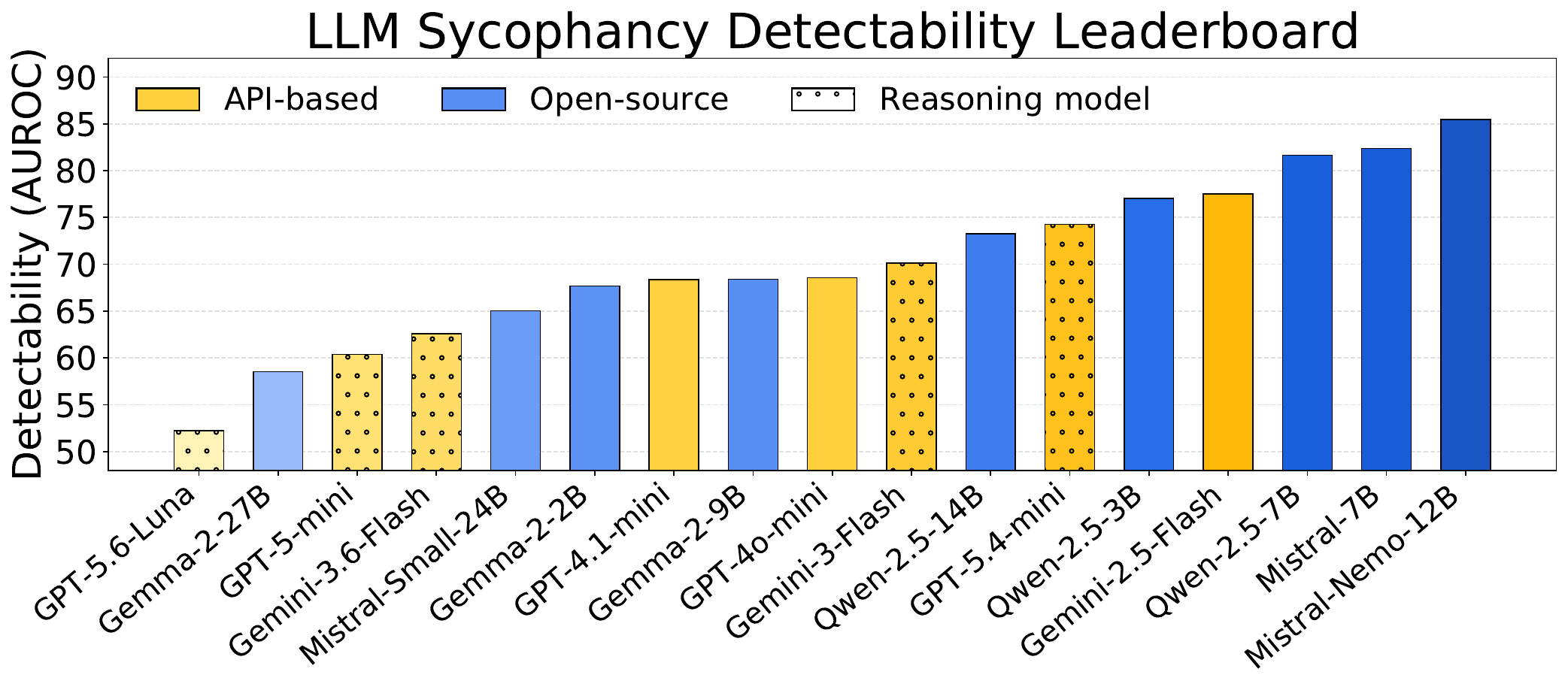}
    \caption{LLMs' PSRS detectability rates from low to high.}
    \label{fig:detectability_rank}
    \vspace{-0.5\baselineskip}
\end{wrapfigure}

Pooling all models by topic category (Figure~\ref{fig:sycophancy_x_model_topic}, right), education at $39\%$, relationships and workplace at $32\%$, and consumer choices at $31\%$ sit at the top, while environment at $7\%$, health at $13\%$, and civic norms at $14\%$ sit at the bottom. We speculate that models hold firm when their anchor reflects a widely shared norm. They tend to be sycophantic when the topic sounds personal (i.e., the user is entitled to their own preference).

\begin{wrapfigure}{r}{0.56\columnwidth}
    \centering
    \vspace{-0.5\baselineskip}
    \includegraphics[width=\linewidth]
    {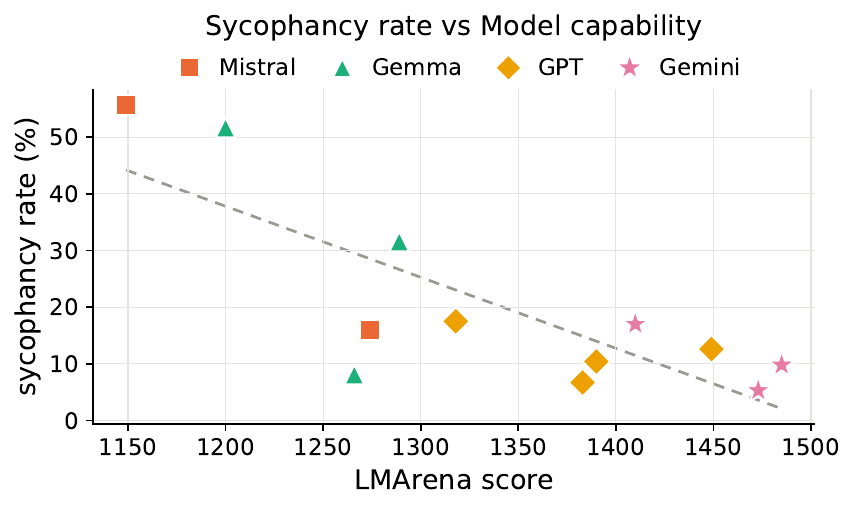}
    \caption{Correlation between models' LMArena scores
    (general capability) and their sycophancy rates.}
    \label{fig:corr_a}
    \vspace{-0.5\baselineskip}
\end{wrapfigure}

\noindent\paragraph{More capable models are less sycophantic.}
Using LMArena overall scores\footnote{\url{https://arena.ai/blog/arena-leaderboard-dataset}} as a proxy for LLMs' general capability, sycophancy rate correlates strongly and negatively with model capability (Figure~\ref{fig:corr_a}). Note that five models are missing from the LMArena leaderboard dataset. So we only present 12 models in Figure~\ref{fig:corr_a}.

\subsection{RQ2. How well can it be detected?}

\noindent\paragraph{Fine-tuned classifiers outperform zero-shot LLM judges.}
\begin{table*}[t]
\centering
\setlength{\tabcolsep}{3pt}
\renewcommand{\arraystretch}{1.43}
\large
\begin{adjustbox}{max width=\textwidth}
\begin{NiceTabular}{l cc cc cc cc cc cc}
\toprule
\rowcolor{headergray}
\Block{2-1}{{\large\textbf{Detector}}}
& \multicolumn{2}{c}{{\large\textbf{Qwen}}}
& \multicolumn{2}{c}{{\large\textbf{Mistral}}}
& \multicolumn{2}{c}{{\large\textbf{Gemma}}}
& \multicolumn{2}{c}{{\large\textbf{GPT}}}
& \multicolumn{2}{c}{{\large\textbf{Gemini}}}
& \multicolumn{2}{c}{{\large\textbf{Avg.}}} \\
\cmidrule(lr){2-3}\cmidrule(lr){4-5}\cmidrule(lr){6-7}\cmidrule(lr){8-9}\cmidrule(lr){10-11}\cmidrule(lr){12-13}
\rowcolor{headergray}
& {\small\textbf{AUC (\%)}} & {\small\textbf{Acc (\%)}}
& {\small\textbf{AUC (\%)}} & {\small\textbf{Acc (\%)}}
& {\small\textbf{AUC (\%)}} & {\small\textbf{Acc (\%)}}
& {\small\textbf{AUC (\%)}} & {\small\textbf{Acc (\%)}}
& {\small\textbf{AUC (\%)}} & {\small\textbf{Acc (\%)}}
& {\small\textbf{AUC (\%)}} & {\small\textbf{Acc (\%)}} \\
\midrule

\Block{1-13}{\textit{\textbf{Statistical detectors}}} \\
\rowcolor{rowgray}
Likelihood$^{\ddagger}$
& 50.9\pmv{1.1} & 50.0\pmv{0.0} & 52.6\pmv{0.7} & 50.0\pmv{0.0}
& 57.5\pmv{0.7} & 50.0\pmv{0.0} & 47.7\pmv{0.8} & 50.0\pmv{0.0}
& 56.9\pmv{0.7} & 50.8\pmv{0.1} & 53.1 & 50.2 \\
LogRank$^{\ddagger}$
& 49.0\pmv{0.9} & 49.7\pmv{0.0} & 54.1\pmv{0.8} & 49.9\pmv{0.1}
& 57.6\pmv{0.7} & 50.0\pmv{0.0} & 48.6\pmv{0.9} & 49.7\pmv{0.3}
& 57.2\pmv{0.7} & 51.8\pmv{0.1} & 53.3 & 50.2 \\
\rowcolor{rowgray}
LRR$^{\ddagger}$
& 49.3\pmv{0.5} & 49.7\pmv{0.0} & 57.3\pmv{0.8} & 50.0\pmv{0.0}
& 56.2\pmv{0.7} & 49.8\pmv{0.0} & 51.2\pmv{0.8} & 50.2\pmv{0.3}
& 55.9\pmv{0.7} & 50.8\pmv{0.4} & 54.0 & 50.1 \\
Entropy$^{\ddagger}$
& 48.1\pmv{0.8} & 50.0\pmv{0.0} & 52.3\pmv{0.5} & 50.2\pmv{0.1}
& 54.2\pmv{1.0} & 50.0\pmv{0.0} & 45.6\pmv{1.0} & 50.0\pmv{0.0}
& 59.3\pmv{1.7} & 51.9\pmv{1.0} & 51.9 & 50.4 \\
\midrule

\Block{1-13}{\textit{\textbf{TF-IDF-based detectors}}} \\
\rowcolor{rowgray}
Logistic regression$^{\dagger}$
& 68.1\pmv{1.2} & 62.5\pmv{0.7} & 62.3\pmv{0.9} & 58.3\pmv{1.3}
& 55.0\pmv{0.8} & 53.8\pmv{0.7} & 57.1\pmv{1.6} & 54.0\pmv{0.9}
& 60.6\pmv{2.4} & 56.8\pmv{1.4} & 60.6 & 57.1 \\
Linear SVM$^{\dagger}$
& 66.5\pmv{1.5} & 61.0\pmv{2.7} & 62.0\pmv{1.1} & 58.1\pmv{1.1}
& 53.9\pmv{0.7} & 53.4\pmv{0.3} & 56.1\pmv{1.5} & 53.2\pmv{1.9}
& 60.2\pmv{2.3} & 52.3\pmv{1.1} & 59.7 & 55.6 \\
\midrule

\Block{1-13}{\textit{\textbf{Fine-tuned transformer-based detectors}}} \\
\rowcolor{rowgray}
RoBERTa-base
& 64.4\pmv{4.0} & 60.2\pmv{3.4} & 80.3\pmv{2.6} & 72.9\pmv{1.9}
& 58.8\pmv{3.1} & 55.0\pmv{1.4} & \textbf{65.0}\pmv{1.8} & 60.5\pmv{1.4}
& 61.0\pmv{4.3} & 55.7\pmv{4.1} & 65.9 & 60.9 \\
RoBERTa-large
& \underline{68.6}\pmv{4.2} & 63.8\pmv{3.3} & \textbf{88.1}\pmv{1.1} & 80.3\pmv{1.7}
& \textbf{69.4}\pmv{4.1} & 62.8\pmv{5.7} & \underline{64.9}\pmv{2.5} & 62.3\pmv{3.1}
& 59.7\pmv{3.7} & 54.2\pmv{1.6} & \textbf{70.1} & 64.7 \\
\rowcolor{rowgray}
BERT-base
& 63.9\pmv{4.3} & 60.2\pmv{4.4} & 70.3\pmv{1.7} & 64.9\pmv{2.1}
& 57.3\pmv{0.9} & 56.1\pmv{2.2} & 53.3\pmv{3.4} & 53.6\pmv{3.7}
& 57.7\pmv{3.8} & 54.3\pmv{1.4} & 60.5 & 57.8 \\
DistilBERT-base
& 56.6\pmv{2.0} & 54.4\pmv{3.3} & 69.7\pmv{1.9} & 65.4\pmv{2.8}
& 56.7\pmv{1.8} & 56.2\pmv{1.8} & 52.4\pmv{2.2} & 52.7\pmv{2.2}
& 58.8\pmv{2.9} & 55.0\pmv{1.7} & 58.8 & 56.7 \\
\rowcolor{rowgray}
ELECTRA-base
& \textbf{69.0}\pmv{4.1} & 63.2\pmv{1.7} & 79.1\pmv{1.2} & 71.1\pmv{1.0}
& 61.9\pmv{3.2} & 58.6\pmv{3.0} & 64.1\pmv{3.3} & 60.2\pmv{2.1}
& \textbf{65.1}\pmv{1.3} & 60.3\pmv{1.4} & \underline{67.8} & 62.7 \\
DeBERTa-v3-base
& 64.3\pmv{4.9} & 59.9\pmv{3.7} & \underline{82.2}\pmv{2.2} & 73.0\pmv{2.5}
& 60.9\pmv{0.9} & 56.9\pmv{1.1} & 58.9\pmv{6.3} & 58.6\pmv{3.6}
& 59.0\pmv{1.9} & 54.5\pmv{1.6} & 65.1 & 60.6 \\
\midrule

\Block{1-13}{\textit{\textbf{Zero-shot LLM judges}}} \\
\rowcolor{rowgray}
Qwen2.5-7B-Instruct$^{\ddagger}$
& 48.2\pmv{1.9} & 51.2\pmv{0.5} & 58.0\pmv{0.5} & 55.5\pmv{0.4}
& \underline{62.9}\pmv{1.2} & 55.7\pmv{1.1} & 64.3\pmv{0.9} & 59.8\pmv{0.3}
& \underline{61.7}\pmv{3.0} & 54.4\pmv{0.4} & 59.0 & 55.3 \\
Llama-3.1-8B-Instruct$^{\ddagger}$
& 50.6\pmv{3.2} & 49.3\pmv{1.7} & 50.1\pmv{1.7} & 51.6\pmv{0.8}
& 51.4\pmv{1.1} & 49.8\pmv{0.2} & 59.1\pmv{3.1} & 54.8\pmv{1.5}
& 56.6\pmv{3.2} & 53.9\pmv{2.0} & 53.6 & 51.9 \\
\bottomrule
\end{NiceTabular}
\end{adjustbox}
\caption{
Detection benchmark on the five model families. Each family column reports AUROC (\%) $\uparrow$ and Accuracy (\%) $\uparrow$, with standard deviations from five runs. Superscripts mark how variance is estimated: (1) $\dagger$ denotes five training topic subsamples and (2) $\ddagger$ denotes five evaluation-topic resamples. Unmarked rows use five training seeds. Chance is 50 for both metrics. \textbf{Bold} and \underline{underline} mark the best and second-best AUROC per family. Accuracy is secondary and is not marked.}
\label{tab:benchmark}
\end{table*}

\begin{wrapfigure}{r}{0.56\columnwidth}
    \centering
    \vspace{-0.5\baselineskip}
    \includegraphics[width=0.975\linewidth]
    {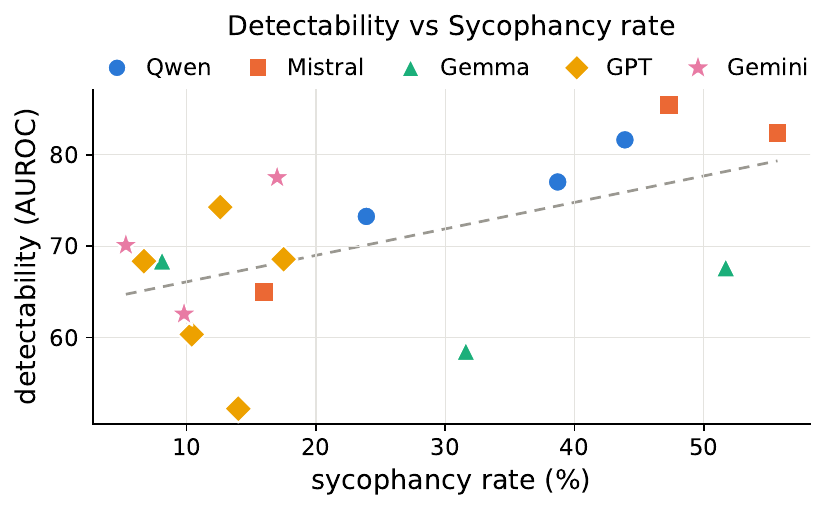}
    \caption{Correlation between models' sycophancy rates
    and their detectability.}
    \label{fig:corr_b}
    \vspace{-0.5\baselineskip}
\end{wrapfigure}

We report the full benchmark in Table~\ref{tab:benchmark}. Fine-tuned transformers lead on every LLM family. RoBERTa-large achieves $88.1\%$ on Mistral and $69.4\%$ on Gemma, ELECTRA-base achieves $69.0\%$ on Qwen and $65.1\%$ on Gemini, and RoBERTa-base achieves $65.0\%$ on GPT. On average, RoBERTa-large performs the strongest at $70.1\%$, followed by ELECTRA-base at $67.8\%$.

The performance of statistical detectors sits between $45.6\%$ and $59.3\%$ AUROC, and around $50\%$ for Accuracy. Zero-shot LLM judges are also near chance on the open-source model families, although Qwen2.5-7B becomes competitive on the two closed families at $64.3\%$ and $61.7\%$. These results indicate that there are \textbf{subtle learnable patterns in the AI response}, since supervised fine-tuned detectors clearly exceed random guessing. Moreover, we observe that prompting LLMs zero-shot for this task is less effective, with averaged accuracy almost at random guessing.

\noindent\paragraph{Detectability differs sharply across models.}
We demonstrate the detectability rankings in Figure~\ref{fig:detectablity_rank}. We can observe that sycophantic responses generated by Mistral-Nemo-12B, Mistral-7B, and Qwen2.5-7B are relatively easy to detect. 
Meanwhile, detecting GPT-5.6-Luna's sycophantic response is challenging, almost at random guess across all baselines. Moreover, we find no clear evidence that sycophantic responses from open-source models are easier to detect.

\noindent\paragraph{Detectability is positively correlated with sycophancy rate.}
Figure~\ref{fig:corr_b} shows that models that flatter more are easier to detect. This observation potentially indicates that sycophantic models may exhibit specific PSRS patterns or templates, which can be learned by detectors. A likely explanation is that more sycophantic models flatter in a more templated way. We also show that there is no clear correlation between detectability and LMArena score (model capability) in the Appendix.

\subsection{RQ3. Does Detection Generalize to Unseen Models?}
New LLMs are released continuously and rapidly, so a deployed detector will encounter models absent from its training data. We evaluate this out-of-distribution (OOD) setting in two steps. We first ask how a detector trained on a single family transfers to another. Then we test whether pooling several source models with a simple domain generalization method improves OOD performance.

\begin{wrapfigure}{r}{0.58\columnwidth}
    \centering
    \vspace{-0.5\baselineskip}
    \includegraphics[width=\linewidth]{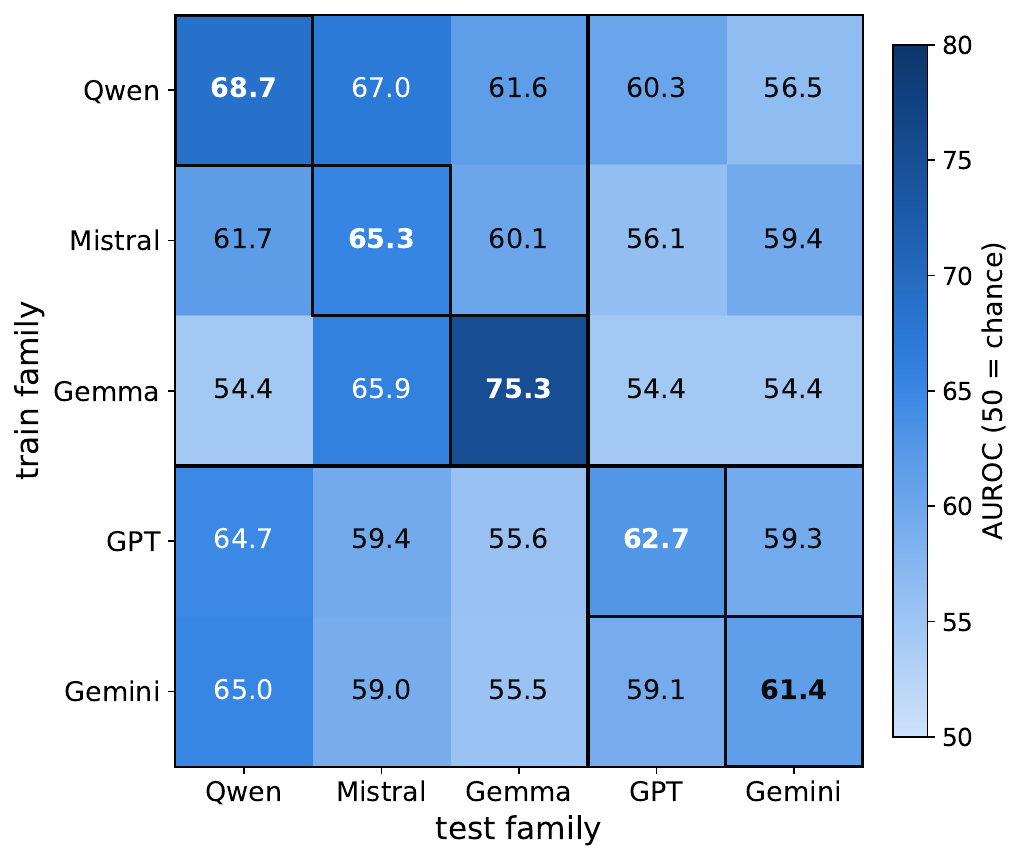}
    \caption{Cross-family PSRS detection performance of RoBERTa-base
    detectors. Rows and columns denote the training and test model
    families, respectively.}
    \label{fig:heatmap5}
    \vspace{-0.5\baselineskip}
\end{wrapfigure}

\noindent\paragraph{Single-source transfer across families is weak.}
As shown in Figure~\ref{fig:heatmap5}, we train RoBERTa-base on one family and test on every family under the balanced protocol. We select 90 topics per family and a global five-fold topic split so that every train and test set is disjoint at the topic level. Because families hold different topic sets (Table~\ref{tab:dataset}), we restrict to topics shared across models for a fair comparison. Averaged over the matrix, staying within a family (i.e., in-distribution) achieves $66.7\%$ AUROC, while crossing families (out-of-distribution) achieves $59.5\%$, showing that PSRS is partly model-specific. The within-family diagonal is strongest for Qwen, Mistral, and Gemma, clearly outperforming their cross-family counterparts. Moreover, the generalization from open- to closed-source model families is relatively weak. For example, Gemma-to-GPT achieves only $54.4\%$ and Mistral-to-GPT $56.1\%$. However, the reverse direction is better, with GPT to Qwen at $64.7\%$ and Gemini to Qwen at $65.0\%$. We can observe that a detector trained on the subtler PSRS of closed-source models can achieve relatively better out-of-distribution detection performance. Meanwhile, detectors trained on the sycophantic data generated by open-source models may overfit. At the fine-grained model level, we find that models within a family may share a similar PSRS pattern, which supports grouping models by family. We include the full cross-model heatmap in the Appendix.

\noindent\paragraph{A simple solution for OOD PSRS detection.}
\begin{table}[t]
\centering
\setlength{\tabcolsep}{6pt}
\renewcommand{\arraystretch}{1.2}
\begin{NiceTabular}{lcc}
\toprule
\rowcolor{headergray}
\textbf{Method} & \textbf{Leave-one-family-out} & \textbf{Open-to-close} \\
\midrule
ERM      & 85.01\pmv{3.65} & 78.23\pmv{1.40} \\
\rowcolor{rowgray}
Mixup    & 85.97\pmv{2.34} & 78.04\pmv{0.70} \\
GroupDRO & \underline{86.20}\pmv{2.70} & \underline{78.63}\pmv{1.79} \\
\rowcolor{rowgray}
CORAL    & 86.17\pmv{2.84} & 78.47\pmv{1.06} \\
\midrule
\textbf{Ours} & \textbf{87.31}\pmv{2.25} & \textbf{79.47}\pmv{1.84} \\
\bottomrule
\end{NiceTabular}
\caption{Cross-model PSRS detection results in AUROC (\%) $\uparrow$.
Leave-one-family-out holds out one open-source model family. Open-to-Close trains
on all nine open-source models and tests on the GPT family. \textbf{Bold} and
\underline{underline} mark the best and second-best result per column.}
\label{tab:dg}
\end{table}
In practice, a detector is usually trained on all available models rather than on a single model family. Thus, we aggregate several source model families and test on a held-out target model family. We treat each source model as a separate environment. 
Our proposed method contains three simple parts. First, a \textbf{contrastive loss} pulls together responses that share a topic and a label but come from different models. This teaches the detector what PSRS looks like rather than how a specific model flatters. Second, a \textbf{variance penalty} keeps the loss even across the source models, so the detector does not rely on any single one. Third, we \textbf{average the model weights over training} for a more stable detector. 
We compare four popular baselines. \textbf{ERM}~\cite{vapnik2013nature} simply aggregates all source data and minimizes the average loss, with no domain-specific term. \textbf{Mixup}~\cite{zhang2017mixup} is a data augmentation method and trains on random blends of two samples and their labels to smooth the decision boundary. \textbf{GroupDRO}~\cite{sagawa2019distributionally} optimizes the worst source domain to make the detector stay robust with domain shifts. \textbf{CORAL}~\cite{sun2016return} aligns the feature statistics across source domains to make the latent representations close to each other.
Table~\ref{tab:dg} reports the task performance for two protocols. \textit{Leave-one-family-out} holds out one of the three open-source model families and trains on the other two. \textit{Open-to-close} trains on all nine open-source models and tests on the five closed-source GPT models. This protocol better mimics the real-world scenario. In practice, open-source models are deployed locally to generate data and train the detector. When detecting in the wild, the detector mostly sees responses from closed-source models, which dominate everyday usage. Aggregating several source model families for training already improves OOD detection performance at the single-source level in Figure~\ref{fig:heatmap5}, and our method improves it further. Our method outperforms all baselines for both protocols. Our method achieves $87.31\%$ on Leave-one-family-out and $79.47\%$ on Open-to-close, respectively. However, the improvements are marginal, and we only present the method as a first step. We aim to demonstrate that there is a lot of room for future work to explore and improve.

\section{Limitations}
In this section, we list limitations in this study as inspirations for future work. First, this paper focuses on a tightly defined form of harmful AI sycophancy --- preference-induced stance reversal sycophancy. Other forms of AI sycophancy such as excessive praise, emotional validation, and factual manipulation are out of scope. Second, in practice, human-AI conversations can be multi-turn and open-ended. So it is promising to systematically explore other types of harmful sycophancy in different conversational scenarios. Third, in our task, the input for the detector only contains an English response, so transfer to other languages, modalities, or multilingual and multimodal cases remains underexplored. Finally, our study collected data from seventeen LLMs from five model families, and our proposed method for cross-modal PSRS detection provides only a small gain. Therefore, we consider that detecting AI sycophancy on unseen models remains an open and important problem.

\section{Conclusion}
We formalize detecting harmful AI sycophancy as a response-level binary classification task in a black-box setting, focusing on preference-induced stance reversal in everyday advice. To study it at scale, we introduce CAP, a controlled framework that first anchors a model's own stance and then tests whether an opposing user preference makes it reverse. Applying CAP to seventeen open- and closed-source LLMs across twelve domains, we collect 290,460 labeled responses. Using this dataset, we conduct extensive experiments guided by three research questions. We conclude our main findings as follows. First. Sycophancy varies widely across models and is more prevalent in less capable ones. Second, fine-tuned detectors consistently outperform zero-shot LLMs in this task. This indicates that there are learnable PSRS patterns in the response text. Therefore, future research could explore how to better extract and leverage these patterns to improve the detection task. Third, we show that the cross-modal PSRS detection has relatively poor performance. Our simple solution can help but does not close the gap significantly. We release our dataset and code to support future work on detecting harmful AI sycophancy.

\bibliographystyle{abbrvnat}
\nobibliography*
\bibliography{custom}

\end{document}